%% file: main.tex
\documentclass[10pt,twocolumn,letterpaper]{article}

\usepackage{cvpr}              
\usepackage[accsupp]{axessibility}

\input{preamble}

\definecolor{cvprblue}{rgb}{0.21,0.49,0.74}
\usepackage[pagebackref,breaklinks,colorlinks,citecolor=cvprblue]{hyperref}
\usepackage{multirow}

\usepackage{soul}

\def\paperID{22} 
\def\confName{CVPR}
\def\confYear{2024}

\title{Learning Relighting and Intrinsic Decomposition in Neural Radiance Fields}

\author{Yixiong Yang$^1$, Shilin Hu$^2$, Haoyu Wu$^2$, Ramon Baldrich$^1$, Dimitris Samaras$^2$, Maria Vanrell$^1$\\
$^1$ Universitat Autonoma de Barcelona, $^2$ Stony Brook University\\
{\tt\small \{yixiong, ramon, maria\}@cvc.uab.cat, \{shilhu, haoyuwu, samaras\}@cs.stonybrook.edu}
}

\begin{document}

\maketitle
\input{sec/0_abstract}    
\input{sec/1_intro}

\clearpage
{
    \small
    \bibliographystyle{ieeenat_fullname}
    \bibliography{main}
}
\input{sec/X_suppl}

\end{document}

%% file: preamble.tex
\usepackage[dvipsnames]{xcolor}

\usepackage{color,xcolor,ulem}



%% file: sec/0_abstract.tex
\begin{abstract}
The task of extracting intrinsic components, such as reflectance and shading, from neural radiance fields is of growing interest. However, current methods largely focus on synthetic scenes and isolated objects, overlooking the complexities of real scenes with backgrounds. To address this gap, our research introduces a method that combines relighting with intrinsic decomposition. By leveraging light variations in scenes to generate pseudo labels, our method provides guidance for intrinsic decomposition without requiring ground truth data. Our method, grounded in physical constraints, ensures robustness across diverse scene types and reduces the reliance on pre-trained models or hand-crafted priors. We validate our method on both synthetic and real-world datasets, achieving convincing results. Furthermore, the applicability of our method to image editing tasks demonstrates promising outcomes.
\end{abstract}

%% file: sec/1_intro.tex
\vspace{-3mm}
\section{Introduction}
\label{sec:intro}

Recent advances in neural rendering have made significant strides in novel view synthesis \cite{mildenhall2021nerf, li2023neuralangelo}, ranging from small objects to large-scale scenes. Concurrently, there has been an exploration towards scene editing \cite{wang2023udcnerf},  such as recoloring \cite{Ye2023IntrinsicNeRF}  and relighting \cite{ling2022shadowneus, zeng2023nrhints}. To facilitate editing, it often becomes necessary to decompose scenes into editable sub-attributes. Within the task of scene decomposition into geometry, reflectance, and illumination using neural rendering, two lines of work are particularly noteworthy: inverse rendering and intrinsic decomposition.

The first approach \cite{Jin2023TensoIR, zhang2022invrender, yang2023sireir, zhang2021nerfactor} integrates inverse rendering with neural rendering methods for scene decomposition. They often employ the BRDF model, such as the simplified Disney BRDF model\cite{burley2012physically}, to model material properties and jointly optimize geometry, BRDF, and environmental lighting. However, inverse rendering presents a highly ill-posed challenge: separating material properties and illumination in images often yields ambiguous results, and tracing light within scenes is computationally intensive. These factors limit inverse rendering to object-specific scenarios. 

The second approach \cite{Ye2023IntrinsicNeRF}, based on intrinsic decomposition\cite{barrow1978recovering}, aims to provide an interpretable representation of a scene (in terms of reflectance and shading) suitable for image editing. It can be considered a simplified variant of inverse rendering, making it more applicable to a broader range of scenarios, including individual objects and more complex scenes with backgrounds. However, despite simplifications over inverse rendering, previous attempts at applying intrinsic decomposition to neural rendering have shown limited success. This motivates our work in this paper. 

Our inspiration is drawn from the idea of using neural rendering to combine relighting and intrinsic decomposition, aiming not only to enhance the quality of intrinsic decomposition but also to expand editing capabilities. Just as experts in mineral identification illuminate specimens from various angles to reveal their features, varying light source positions are essential for uncovering a scene's intrinsic details. In fact, the connection between relighting and intrinsic decomposition has been discussed in previous works on 2D images \cite{BigTimeLi18,lettry2018unsupervised}, but it has yet to be explored in neural rendering. Additionally, the field of neural rendering has significantly explored relighting \cite{zeng2023nrhints, rudnev2022nerfosr}. While IntrinsicNeRF \cite{Ye2023IntrinsicNeRF} has pioneered the integration of intrinsic decomposition within NeRF, they have not utilized relighting or fully leveraged the 3D information available through neural rendering. Instead, we focus on physics-based constraints to enhance the intrinsic decomposition performance.

In this paper, we propose a two-stage method. In the first stage, we train a neural implicit radiance representation to enable novel view synthesis and relighting. Based on the results of this stage, we calculate normals and light visibility for each training image, which allows us to develop a method for generating pseudo labels for reflectance and shading. In the second stage, we treat reflectance and shading as continuous functions parameterized by Multi-Layer Perceptrons (MLPs). During training, we apply constraints based on physical principles and our pseudo labels. Notably, our approach does not depend on any pre-trained models or ground truth data for intrinsic decomposition, yet achieves convincing results, as shown in \cref{teaser}. Our contributions are summarized as follows:
\begin{itemize}
\item We propose a method that integrates relighting with intrinsic decomposition, allowing for novel view synthesis, lighting condition altering, and reflectance editing.

\item We propose a method to generate pseudo labels for reflectance and shading through neural fields that integrate multiple lighting conditions.

\item Our method, applied to NeRF scenes, operates free from data-driven priors. It factorizes the scene into reflectance, shading, and a residual component, proving effective even in the presence of strong shadows.
\end{itemize}

\section{Related Work}

\noindent\textbf{{Intrinsic decomposition.}}
Intrinsic decomposition is a classical challenge in computer vision \cite{barrow1978recovering}, with much of the previous research focused on the 2D image\cite{dasPIENet, careagaIntrinsic, barron2014shape, li2018cgintrinsics}. A key difficulty in this area is the scarcity of real datasets, which need complicated and extensive annotation. This limitation has spurred interest in semi-supervised and unsupervised techniques\cite{BigTimeLi18,lettry2018unsupervised, liu2020unsupervised}. IntrinsicNeRF \cite{Ye2023IntrinsicNeRF} has been a pioneer in applying intrinsic decomposition to neural rendering. Similar to previous unsupervised methods in 2D, it utilizes hand-crafted constraints, including chromaticity and semantic constraints, for guidance. However, these constraints do not accurately reflect physical principles and often fall short in complex scenarios. Our approach leans on 3D information and physical constraints (e.g., variations in illumination) to achieve superior results.

\noindent\textbf{{Relighting.}}
Relighting has recently garnered attention from various perspectives within the field \cite{einabadi2021deep}. Data-driven approaches have been explored, with research focusing on portrait scenes  \cite{nestmeyer2020learning,sun2019single,zhou2019deep,pandey2021total, hou2022face} and extending to more complex scenarios 
 \cite{murmann2019dataset, helou2020aim, puthussery2020wdrn, wang2020deep, el2021ntire}. Kocsis et al. \cite{kocsis2024lightit} have also investigated lighting control within diffusion models, enabling the generation of scenes under varying lighting conditions. Meanwhile, relighting has also received widespread attention within the field of neural rendering \cite{srinivasan2021nerv, gao2020deferred, zeng2023nrhints, Toschi_2023_CVPR}, achieving impressive relighting outcomes within individual scenes.

\vspace{-1mm}
\section{Method}

Under the Lambertian assumption, images can be decomposed into reflectance and shading components \cite{barrow1978recovering,careagaIntrinsic,fan2018revisiting}. However, real-world scenes often require a residual term to account for discrepancies \cite{garces2022survey, Ye2023IntrinsicNeRF}. Thus, we model intrinsic decomposition as follows:
\vspace{-1mm}
\begin{equation}
I(i, j)=R(i, j) \odot S(i, j) + Re(i, j)
\label{eq:Intrinsic decomposition}
\end{equation}
\noindent where $R$, $S$ and $Re$ denote Reflectance, Shading and Residual, respectively. 

\begin{figure*}[t]
  \centering
   \includegraphics[width=0.87\linewidth]{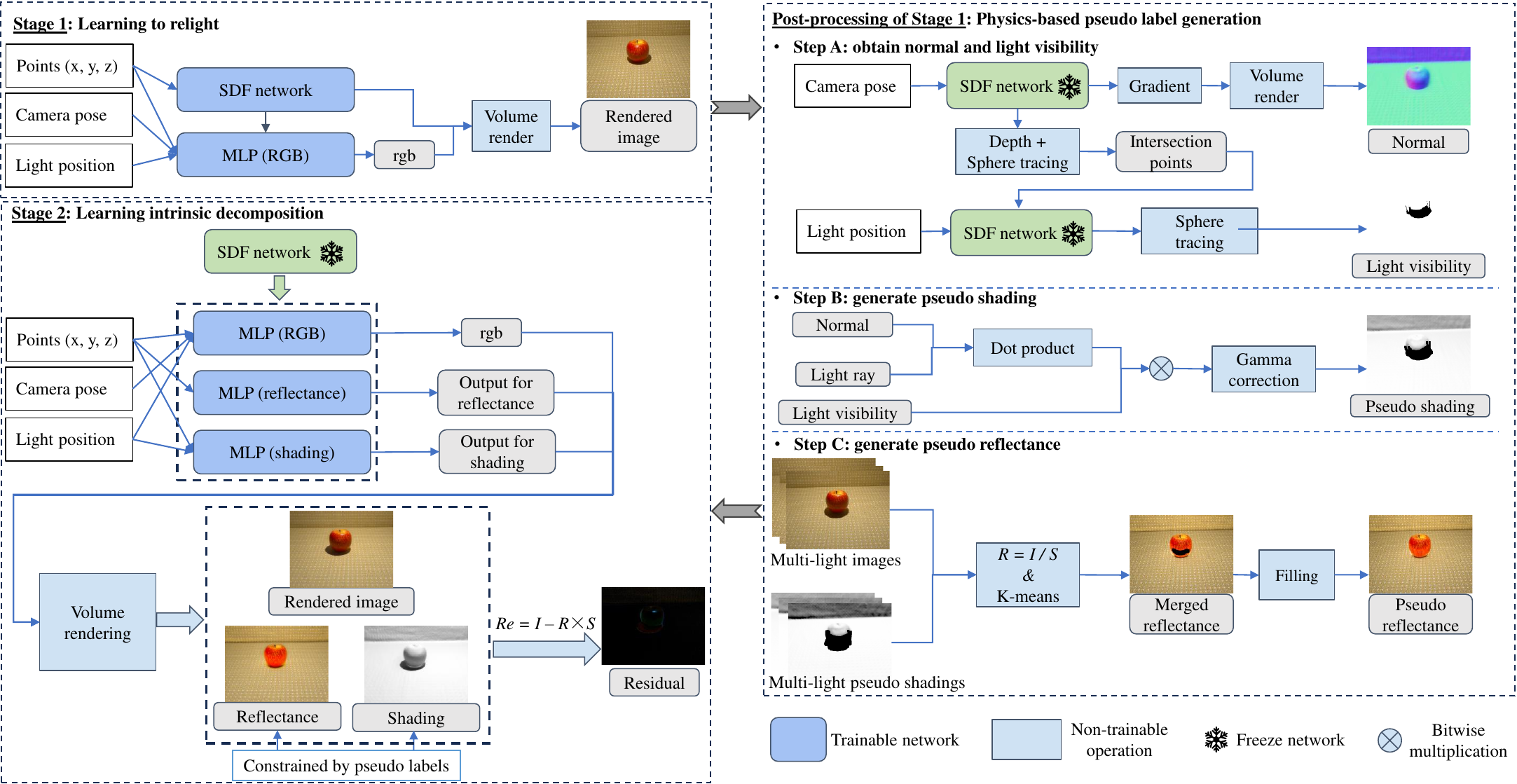}
   \vspace{-2mm}
   \caption{Method Framework: Stage 1 involves learning the neural field with relighting (top left). Post-processing and generating pseudo labels (right). In Stage 2, the learning process continues to learn intrinsic decomposition based on the model trained in Stage 1 and the pseudo labels (bottom left).}
   \label{Method Framework}
   \vspace{-6mm}
\end{figure*}

Our method extends implicit neural representation for relighting and intrinsic decomposition. 
We propose a two-stage approach, illustrated in \cref{Method Framework}.
In the first stage, we train our model to represent scenes under varying camera positions and lighting conditions, enabling novel view synthesis and relighting.
We then apply three steps to generate pseudo labels for reflectance and shading.
In the second stage, we expand the model to decompose intrinsics using these pseudo labels as constraints.
Our proposed model achieves novel view synthesis, relighting, and intrinsic decomposition simultaneously.

\vspace{-1mm}
\subsection{Stage 1: Learning to Relight}

We use 3D hash grids \cite{mueller2022instant, li2023neuralangelo} to represent geometry and a small MLP to model color, which accepts the light position as an input. We illustrate Stage 1 in \cref{Method Framework} (top-left), with formulas as follows:
\begin{equation}
sdf = f(\mathbf{x}), \quad \mathbf{c} = \textrm{MLP}_{color}(\mathbf{x}, \mathbf{d}, \mathbf{l}, \mathbf{feat})
\label{eq: stage 1}
\end{equation}
where $f(\cdot)$ is the geometry network that predicts Signed Distance Function and $\textrm{MLP}_{color}(\cdot)$ is the color network. $\mathbf{x}$ is the spatial position, $\mathbf{d}$ is the view direction, $\mathbf{l}$ is the light position, and $\mathbf{feat}$ is the feature from SDF network.
Following \cite{li2023neuralangelo}, the loss for Stage 1 is:
\begin{equation}
\mathcal{L}_{S1} = \mathcal{L}_{\mathrm{RGB}} + w_{\text{eik}} \mathcal{L}_{\text{eik}} + w_{\text{curv}} \mathcal{L}_{\text{curv}}
\end{equation}
where $\mathcal{L}_{\mathrm{RGB}}$ is the loss of the rendered image, $\mathcal{L}_{\text{eik}}$ represents the Eikonal loss \cite{icml2020_2086}, and $\mathcal{L}_{\text{curv}}$ is the curvature loss. The terms $w_{\text{eik}}$ and $w_{\text{curv}}$ are the corresponding weights.

\subsection{Physics-based Pseudo Label Generation}

Our proposed post-processing aims to generate pseudo labels for reflectance and shading in three steps, as illustrated in \cref{Method Framework} (right).
Based on the physics modeling of image formation, we start with generating pseudo shading by the normal and light visibility. We then generate pseudo reflectance using multiple images and shadings under different illumination.
Details can be found in the supplementary.

\noindent \textbf{Step A.} 
The normals are derived from the SDF network. The geometry network also provides depth information which is used to estimate the intersection points in conjunction with sphere tracing\cite{chen2022tracing}. Light visibility, which indicates whether a point is directly illuminated, is obtained by sphere tracing based on the light position and intersection points.

\noindent \textbf{Step B.} The generation of pseudo-shading follows the formula,
$S^* = ((\Vec{N} \cdot \Vec{L}) \cdot V)^\gamma$,
where the optimal shading $S^*$ is the multiplication of the light visibility $V$ and the dot product of the normal $\Vec{N}$ and the light ray $\Vec{L}$.
$(\cdot)^\gamma$ represents for gamma correction.
This correction is essential because the human eye's perception of brightness is not linear. Thus, we apply it to accommodate the perceptual effect, yielding to our pseudo shading.

\noindent \textbf{Step C.} 
Our method infers pseudo reflectance from pseudo shading using $R=I/S$. For pseudo labels, this calculation is only applicable in the case of direct illumination. We utilize various lighting conditions to obtain different reflectance values; and by employing K-means \cite{scikit-learn} along with the confidence related to pseudo shading, we merge to form the most probable reflectance map. Areas lacking direct illumination are filled using a strategy considering pixel distance, normals, and RGB colors, resulting in the final pseudo reflectance.

\setlength{\tabcolsep}{3pt}
\begin{table*}[!t]
\centering
\begin{tabular}{ccccccccccccc}
\hline
              & \multicolumn{3}{c}{Hotdog (Reflectance)} & \multicolumn{3}{c}{Hotdog (Shading)} & \multicolumn{3}{c}{Lego (Reflectance)} & \multicolumn{3}{c}{Lego (Shading)} \\
              & PSNR$\uparrow$  & SSIM$\uparrow$   & LPIPS$\downarrow$  & PSNR$\uparrow$  & SSIM$\uparrow$   & LPIPS$\downarrow$  & PSNR$\uparrow$  & SSIM$\uparrow$   & LPIPS$\downarrow$  & PSNR$\uparrow$  & SSIM$\uparrow$   & LPIPS$\downarrow$  \\\hline
PIE-Net\cite{dasPIENet}       & 22.21 & 0.9208 & 0.0843 & 22.92 & 0.9291 & 0.1091 & 20.70 & 0.8902 & 0.1127 & 21.36 & 0.9016 & 0.1180 \\
Careaga \etal\cite{careagaIntrinsic}       & 19.67 & 0.9177 & 0.0894 & 18.94 & 0.9346 & 0.0946 & 17.78 & 0.8801 & 0.1123 & 19.63 & 0.9032 & 0.1068 \\
IntrinsicNeRF*\cite{Ye2023IntrinsicNeRF} & 25.62 & 0.9620 & 0.0967 & -     & -      & -      & 19.00 & 0.9046 & 0.1288 & -     & -      & -      \\
Ours     & \textbf{28.87} & \textbf{0.9642} & \textbf{0.0381} & \textbf{27.68} & \textbf{0.9666} & \textbf{0.0330} & \textbf{26.88} & \textbf{0.9472} & \textbf{0.0432} & \textbf{24.45} & \textbf{0.9417} & \textbf{0.0520} \\
\hline
\end{tabular}
  \vspace{-3mm}
  \caption{Quantitative results on the NeRF \cite{mildenhall2021nerf} dataset.}
  \label{tab:Quantitative results on the NeRF dataset.}
  \vspace{-3mm}
\end{table*}

\begin{figure*}[t]
  \centering
   \includegraphics[width=1.0\linewidth]{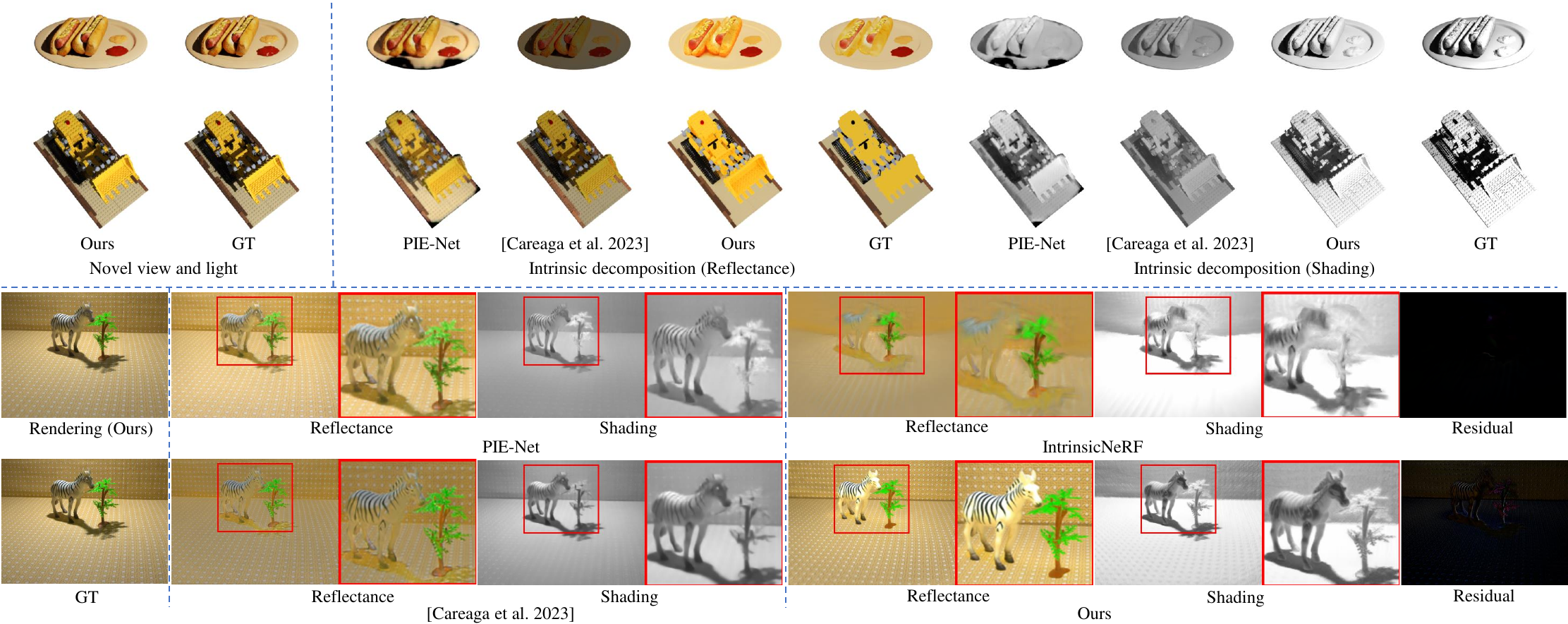}
   \vspace{-8mm}
   \caption{Qualitative comparisons on the NeRF \cite{mildenhall2021nerf} (the first two rows) and ReNe \cite{Toschi_2023_CVPR} datasets (the last two rows).}
   \label{Compare with other methods.}
   \vspace{-6mm}
\end{figure*}

\vspace{-1mm}
\subsection{Stage 2: Learning Intrinsic Decomposition}
As illustrated in \cref{Method Framework} (bottom-left), we jointly learn the relighting and intrinsic decomposition. Expanding the model from Stage 1, we add two extra MLPs dedicated to generating reflectance and shading outputs, while the geometry network is frozen. Note that, while all MLPs receive SDF feature inputs, the RGB color MLP accepts spatial points, camera pose, and light positions as input, the reflectance MLP only receives spatial points, and the shading MLP takes spatial points and light positions.

After volume rendering, we obtain RGB images, along with reflectance and shading. Subsequently, the residual is derived from \cref{eq:Intrinsic decomposition}. 
During training, the pseudo labels are used to impose constraints on reflectance and shading. 
\begin{equation}
L_{intrinsic}=W_R \cdot \|\hat{R}-R^*\|_1 + W_S \cdot \|\hat{S}-S^*\|_1
\label{eq: intrinsic loss}
\end{equation}
where $\hat{R}$ and $\hat{S}$ represent the predicted reflectance and shading, respectively, and $R^*$ and $S^*$ are their corresponding pseudo labels. 
$W_R$ and $W_S$ represent weight maps for reflectance and shading, derived during pseudo label generation.
As demonstrated in \cite{Ye2023IntrinsicNeRF}, the diffuse components dominate the scene, so it is crucial to prevent the training from converging to undesirable local minima ($R=0, S=0, Re=I$). 
Therefore, we introduce a regularization term, $L_{reg}=\|\hat{Re}\|_1$, to ensure that the image is primarily recovered through $R$ and $S$. Finally, the Stage 2 loss is:
\begin{equation}
\mathcal{L}_{S2} = \mathcal{L}_{\mathrm{RGB}} + w_{\text{intrinsic}}L_{intrinsic} + w_{\text{reg}}L_{reg}
\end{equation}

\vspace{-1mm}
\section{Experiments}
\vspace{-1mm}
We conduct experiments on both the NeRF \cite{mildenhall2021nerf} (synthetic) and the ReNe \cite{Toschi_2023_CVPR} (real) datasets.
Detailed setup can be found in the supplementary.
We compare our method with traditional learning-based methods (PIE-Net \cite{dasPIENet} and Careaga \etal \cite{careagaIntrinsic}) and the state-of-the-art neural rendering approach (IntrinsicNeRF \cite{Ye2023IntrinsicNeRF}).

\cref{tab:Quantitative results on the NeRF dataset.} displays our method's quantitative results compared with other methods on the NeRF dataset. Since IntrinsicNeRF struggles with datasets that have lighting variations, we use the numbers from the original publication for comparison. For both the Hotdog and Lego scenes, our approach surpasses others in terms of both reflectance and shading across all metrics.

\cref{Compare with other methods.} presents the qualitative comparison of our method against others on both the synthetic NeRF dataset and the real-world ReNe dataset. On the NeRF dataset, we first showcase the outcomes of our synthesized novel views and lighting conditions on the left, demonstrating results closely aligned with the GT. Then, we display the results of intrinsic decomposition compared to other approaches. 
It is evident that our results are quite convincing and outperform those of others, with almost no lingering cast shadows in the reflectance.
The latter part shows results from the challenging Rene dataset, characterized by real scenes with backgrounds. Our rendering effects, displayed on the left, closely approximate the GT. Moreover, our method is the only one that achieves credible results in intrinsic decomposition. In terms of reflectance, the object's texture edges are sharp, the colors are vibrant, and shadows are accurately eliminated. In contrast, the results from PIE-Net\cite{dasPIENet} and Careaga \etal\cite{careagaIntrinsic} are blurry and fail to remove shadows correctly. The other neural rendering method, IntrinsicNeRF \cite{Ye2023IntrinsicNeRF}, also fails to achieve correct decomposition, primarily attributed to the failure in distinguishing intrinsic components and also the difficulty in scene reconstruction. 
\vspace{-2mm}
\section{Conclusion}
\vspace{-2mm}
We introduce a neural rendering method that learns relighting and intrinsic decomposition from multi-view images with varying lighting without the intrinsic GT. This approach supports the creation of new views, relighting, and decomposition simultaneously, serving as a versatile tool for editing tasks like reflectance and shading adjustments.
Our tests on both synthetic and real-world datasets validate our method's effectiveness.
This method, grounded in basic physical concepts rather than predefined priors, shows promise for more complex scene analyses.
In the future, we aim to extend our experiments to explore a more comprehensive set of scenes.

\noindent \textbf{Acknowledgement:} Thanks to Hassan Ahmed Sial for his assistance in generating the synthetic scenes.
YY, MV and RB were supported by Grant PID2021-128178OB-I00 funded by MCIN/AEI/10.13039/501100011033, and Generalitat de Catalunya 2021SGR01499. YY is supported by China Scholarship Council.

%% file: sec/X_suppl.tex
\clearpage
\setcounter{page}{1}
\maketitlesupplementary

\noindent In this supplementary material, we present the following:
\begin{enumerate}
    \item More detailed procedures for generating pseudo labels.
    \item Specifications of experimental settings.
    \item Additional qualitative results.
\end{enumerate}

\section{Pseudo Label Generation}

Here, we elaborate on the post-processing steps (\cref{post processing}) in the main paper \textbf{Sec.} 3.2.
It starts with generating pseudo shading based on Lambertian reflection principles.
Under the assumption that the light intensity and color remain constant, shading can be approximated by the dot product between the normal ray and the light ray. 
The light ray encompasses both direct/indirect illumination and necessitates light visibility to account for occlusion effects.

\subsection{Step A: obtain normal and light visibility} 
The normals are derived from the SDF network. The geometry network also provides depth information which is used to estimate the intersection points in conjunction with sphere tracing\cite{chen2022tracing}. Light visibility, which indicates whether a point is directly illuminated, is obtained by sphere tracing based on the light position and intersection points.

\subsection{Step B: generate pseudo shading} 
The generation of pseudo-shading follows the formula,
\begin{equation}
    S^\prime = ((\Vec{N} \cdot \Vec{L}) \cdot V)^\gamma
\end{equation}
where the optimal shading $S^\prime$ is the multiplication of the light visibility $V$ and the dot product of the normal $\Vec{N}$ and the light ray $\Vec{L}$.
$(\cdot)^\gamma$ represents for gamma correction.
This correction is crucial because the human eye's perception of brightness is not linear. Most images we see have undergone gamma correction to accommodate this perceptual effect. Therefore, calculating shading also necessitates gamma correction, yielding to our defined pseudo shading.

\begin{figure}[t]
  \centering
   \includegraphics[width=1.0\linewidth]{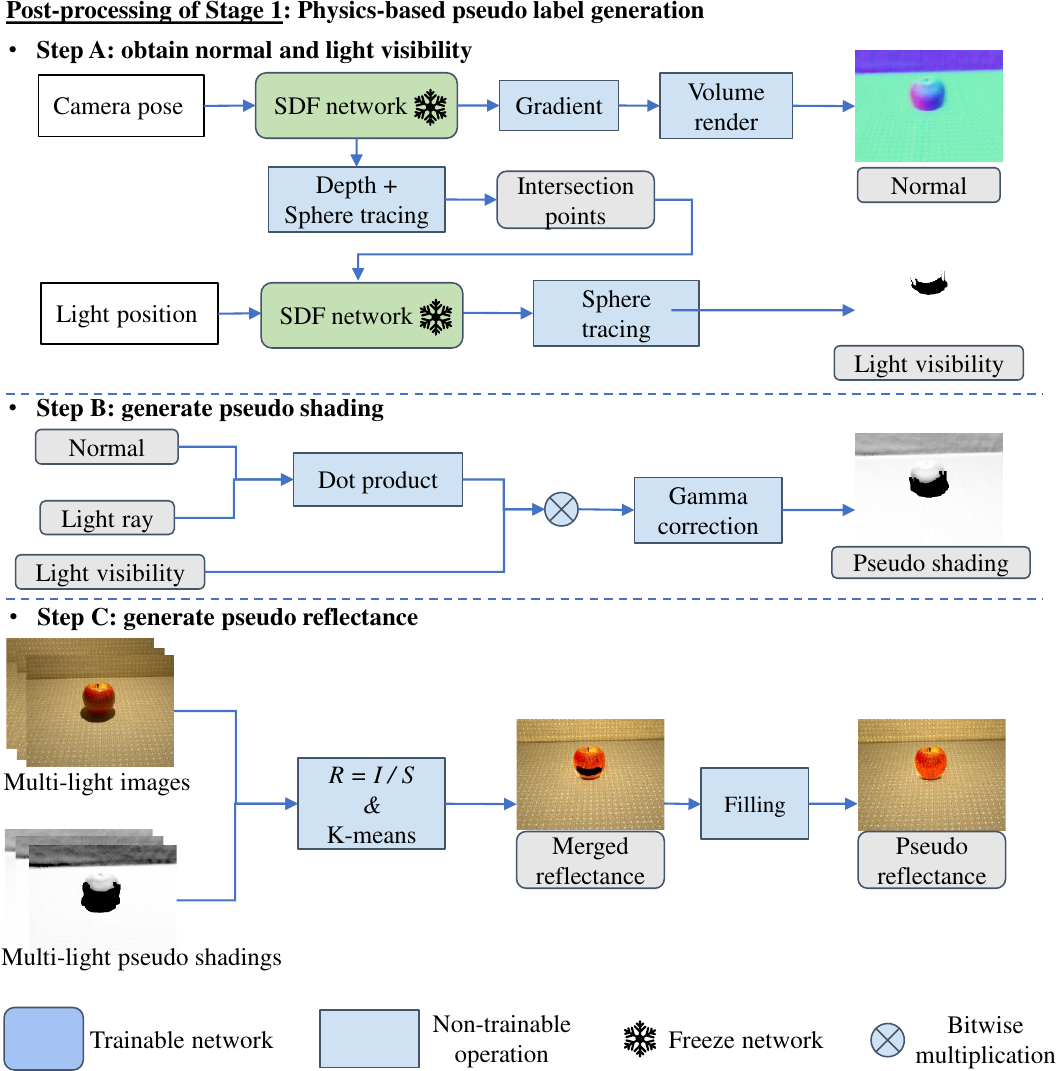}
   \caption{Post-processing and generating pseudo labels.}
   \label{post processing}
   \vspace{-5mm}
\end{figure}

\subsection{Step C: generate pseudo reflectance} 
This step entails inferring the most probable pseudo reflectance from the pseudo shading, principally based on the equation $R=I/S$. The approach has two main points that should be noted here. 

First, the current pseudo shading only considers direct light. As seen in previous papers \cite{zhang2022invrender, yang2023sireir, Jin2023TensoIR}, solving for indirect light is a complex and computationally expensive process. Our novel approach leverages the trained model to generate multiple versions of images under different lighting conditions, each accompanied by respective pseudo shadings. As direct light strengthens on a pixel, the influence of indirect light diminishes, making the reflectance derived from higher pseudo shading values more reliable. We compare the outcomes under multiple lighting conditions and synthesize the most credible reflectance for each pixel based on the intensity of pseudo shading.

Second, the residual term includes specularity and other effects that are not considered in $R=I/S$. Specular highlights, which have high pseudo shading values, do not reflect the object color but rather the light source color (e.g., white reflections). By analyzing different lighting conditions, where highlights typically vanish except under specific angles, we can deduce the object color by selecting the most common reflectance outcomes. 

Our implementation employs the K-means algorithm, incorporating the weights of pseudo shading. This approach allows us to achieve a merged reflectance under varied lighting conditions, as shown in the intermediate result at the bottom in \cref{post processing}. However, some regions within the merged reflectance may appear vacant due to the absence of direct illumination in all lighting conditions. So, we address these areas with a filling strategy. This strategy specifically considers the distance between void and non-void pixels, their normals, and their colors in the RGB image, thereby achieving the final pseudo reflectance.

Additionally, we compute weight maps $W_R$ and $W_S$ for both pseudo reflectance and pseudo shading based on the edges of pseudo shading and visibility. Areas with higher pseudo shading values, or those further from visibility edges (where visibility calculations may be prone to errors), exhibit greater credibility in their pseudo labels; conversely, areas closer to visibility edges or with lower pseudo shading values are deemed less reliable.

\section{Experimental Settings}
\noindent \textbf{Datasets.}
To validate our approach, we conduct experiments on both synthetic and real-world datasets.

For the synthetic dataset, models are obtained from NeRF \cite{mildenhall2021nerf}, with lighting configurations borrowed from Zeng et al. \cite{zeng2023nrhints}. To facilitate quantitative analysis, GT for reflectance, shading, and residuals are rendered in Blender. Each scene comprises 500 images for training, 100 for validation, and 100 for testing, including intrinsic components for each image. Importantly, adhering to the configurations in \cite{zeng2023nrhints}, the settings for lighting and camera poses are managed independently.

The real dataset we use is the ReNe dataset \cite{Toschi_2023_CVPR}, where lighting and camera poses are grid-sampled. This dataset features 2000 images across scenes, captured from 50 different viewpoints under 40 lighting conditions. Following their dataset split, we use 1628 images (44 camera poses $\times$ 37 light positions) for training.

Additionally, given that the settings of lights and cameras are dependent on the former one and grid-sampled in the latter, our proposed method is designed to accommodate both configurations.

\noindent \textbf{Metrics.} To evaluate the comparison between predicted images and ground truth (GT), we employ the following metrics: Peak Signal-to-Noise Ratio (PSNR), Structural Similarity Index (SSIM) \cite{wang2004image}, and Learned Perceptual Image Patch Similarity (LPIPS) \cite{zhang2018unreasonable}.

\noindent \textbf{Implementation details.} Our model's hyperparameters include a batch size of 2048 and each stage was trained for 500k iterations. We implemented the model in PyTorch and used the AdamW \cite{loshchilov2018decoupled} optimizer with a learning rate of $1e^{-3}$ for optimization. The experiments can be conducted on a single Nvidia RTX 3090 or A40 GPU. The weights of losses, $w_{\text{eik}}$, $w_{\text{curv}}$, $w_{\text{intrinsic}}$, $w_{\text{reg}}$ are set to $0.1$, $5e^{-4}$, $1.0$, and $1.0$, respectively. 

\section{Additional qualitative results}
\begin{figure*}[!t]
  \centering
   \includegraphics[width=0.95\linewidth]{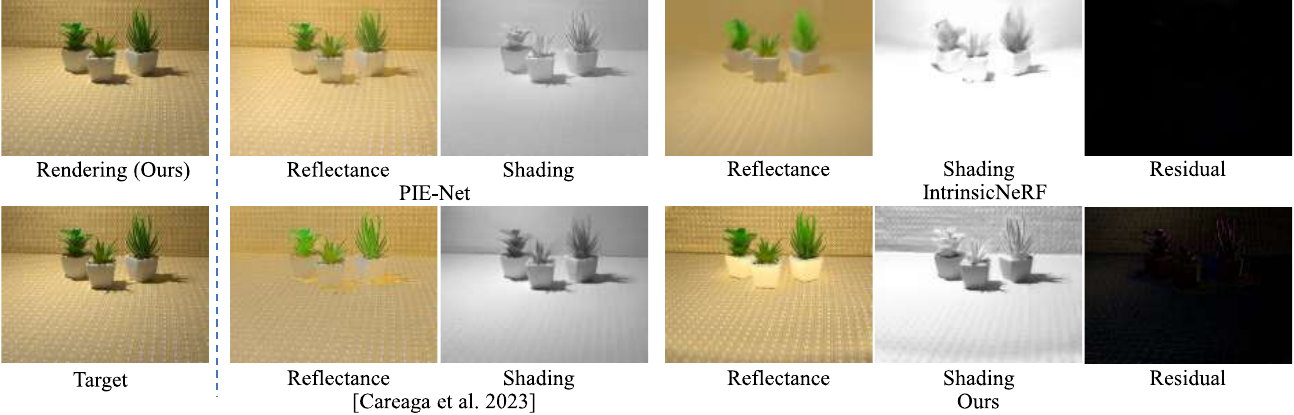}
   \caption{Compare with other methods on the ReNe dataset. }
   \label{Compare with other methods. supplementary}
\end{figure*}

\begin{figure*}[!t]
  \centering
  \includegraphics[width=0.95\textwidth]{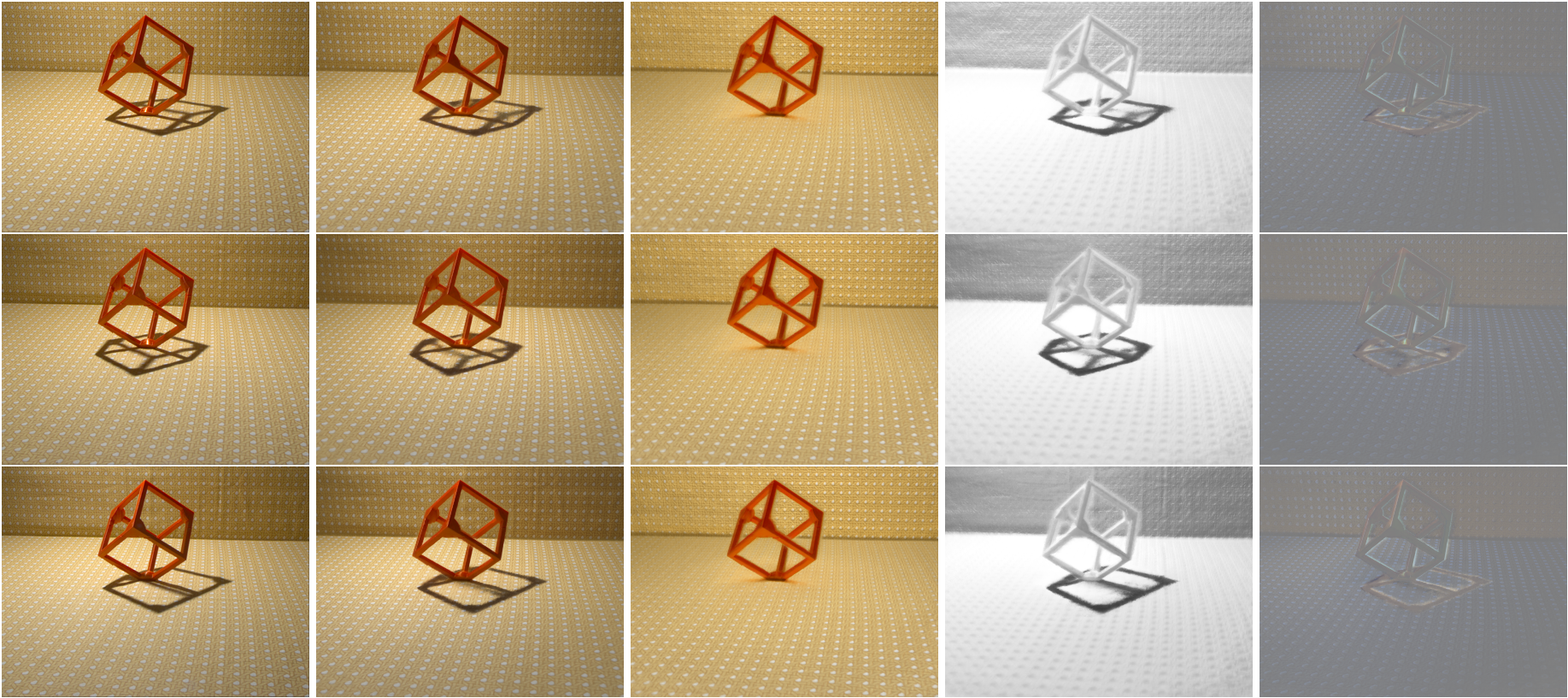}
  \centering
  \begin{minipage}[b]{0.19\linewidth} \centering
   GT  \end{minipage}
   \begin{minipage}[b]{0.19\linewidth} \centering
   Rendering  \end{minipage}
   \begin{minipage}[b]{0.19\linewidth} \centering
   Reflectance  \end{minipage}
   \begin{minipage}[b]{0.19\linewidth} \centering
   Shading  \end{minipage}
   \begin{minipage}[b]{0.19\linewidth} \centering
   Residual  \end{minipage}
  \caption{Qualitative results on the ReNe dataset (Cube). }
  \label{fig: Rene 1}
\end{figure*}

\begin{figure*}[!t]
  \centering
  \includegraphics[width=0.95\textwidth]{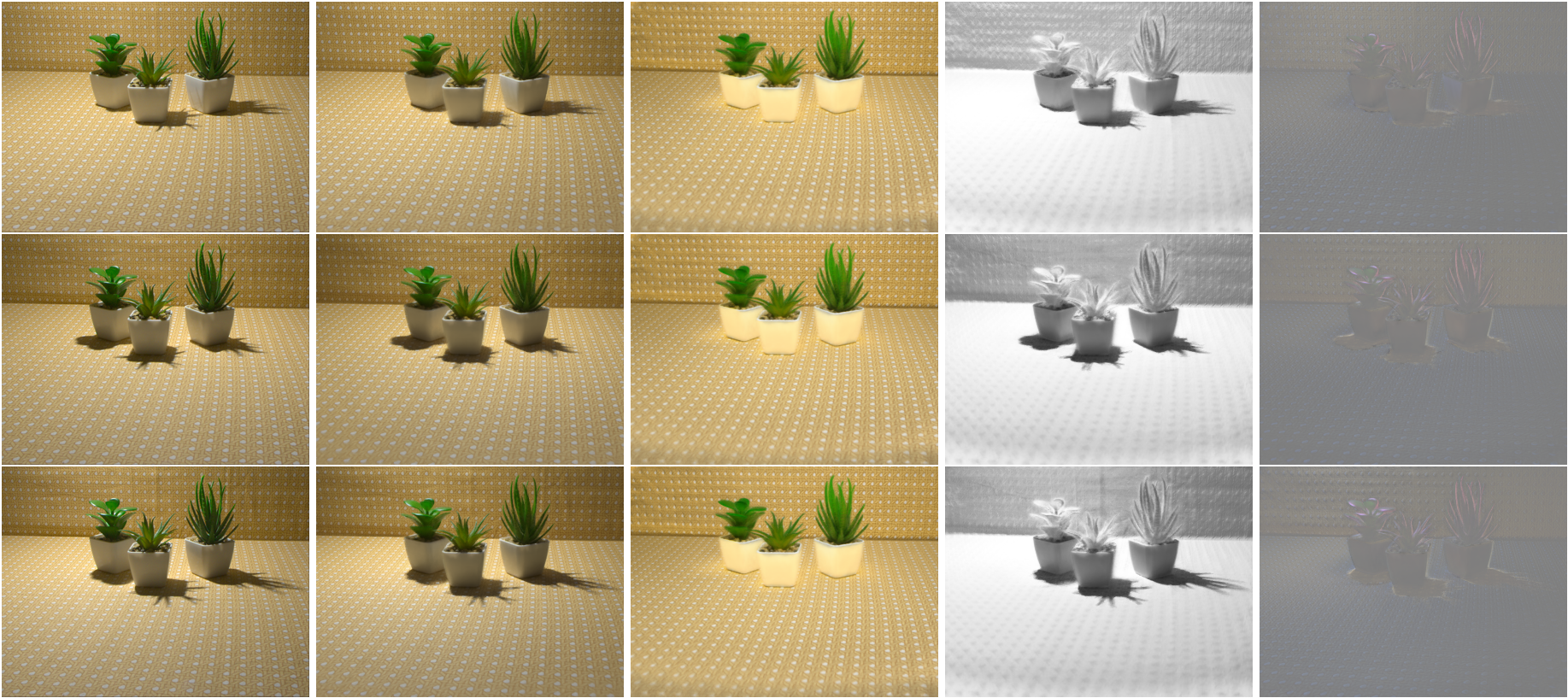}
  \centering
  \begin{minipage}[b]{0.19\linewidth} \centering
   GT  \end{minipage}
   \begin{minipage}[b]{0.19\linewidth} \centering
   Rendering  \end{minipage}
   \begin{minipage}[b]{0.19\linewidth} \centering
   Reflectance  \end{minipage}
   \begin{minipage}[b]{0.19\linewidth} \centering
   Shading  \end{minipage}
   \begin{minipage}[b]{0.19\linewidth} \centering
   Residual  \end{minipage}
  \caption{Qualitative results on the ReNe dataset (Garden). }
  \label{fig: Rene 2}
\end{figure*}

\begin{figure*}[!t]
  \centering
  \includegraphics[width=0.95\textwidth]{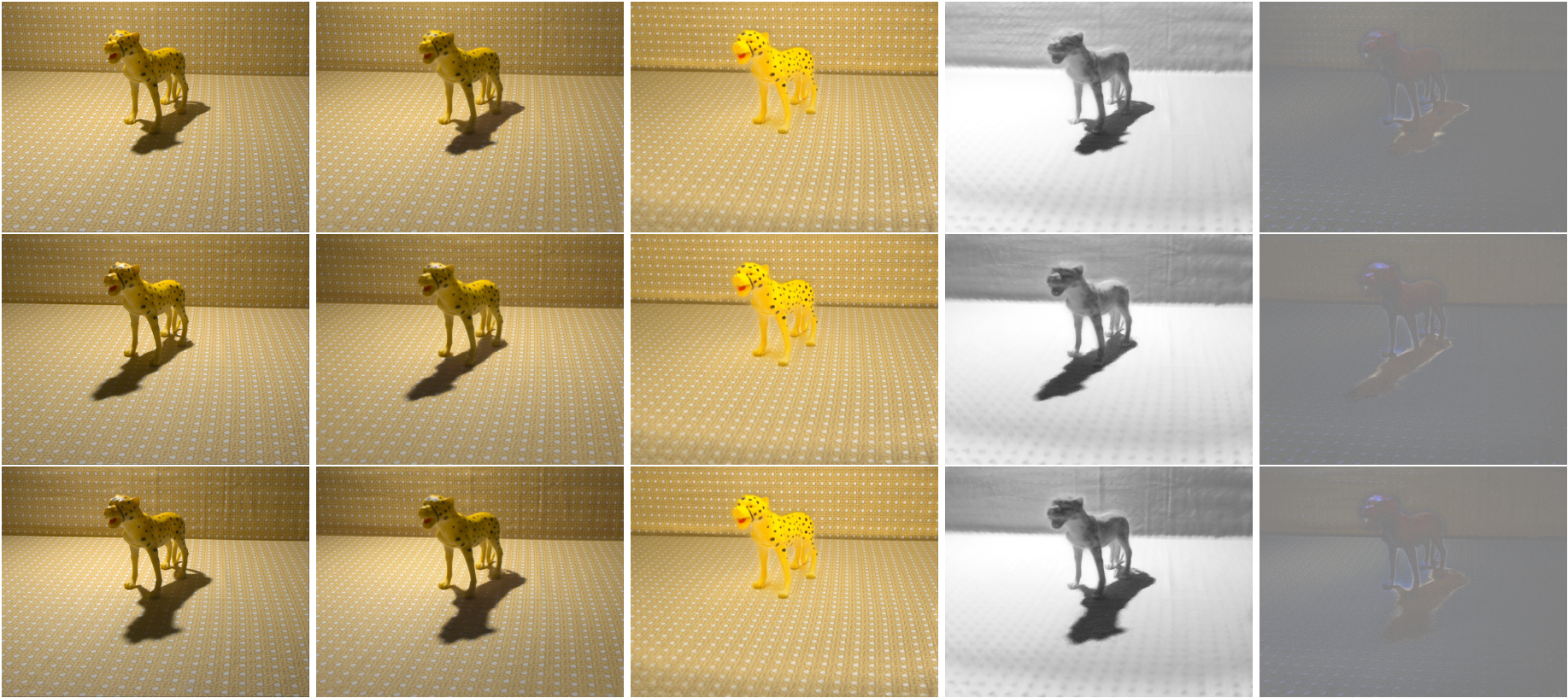}
  \centering
  \begin{minipage}[b]{0.19\linewidth} \centering
   GT  \end{minipage}
   \begin{minipage}[b]{0.19\linewidth} \centering
   Rendering  \end{minipage}
   \begin{minipage}[b]{0.19\linewidth} \centering
   Reflectance  \end{minipage}
   \begin{minipage}[b]{0.19\linewidth} \centering
   Shading  \end{minipage}
   \begin{minipage}[b]{0.19\linewidth} \centering
   Residual  \end{minipage}
  \caption{Qualitative results on the ReNe dataset (Cheetah). }
  \label{fig: Rene 3}
\end{figure*}

\begin{figure*}[!t]
  \centering
  \includegraphics[width=0.95\textwidth]{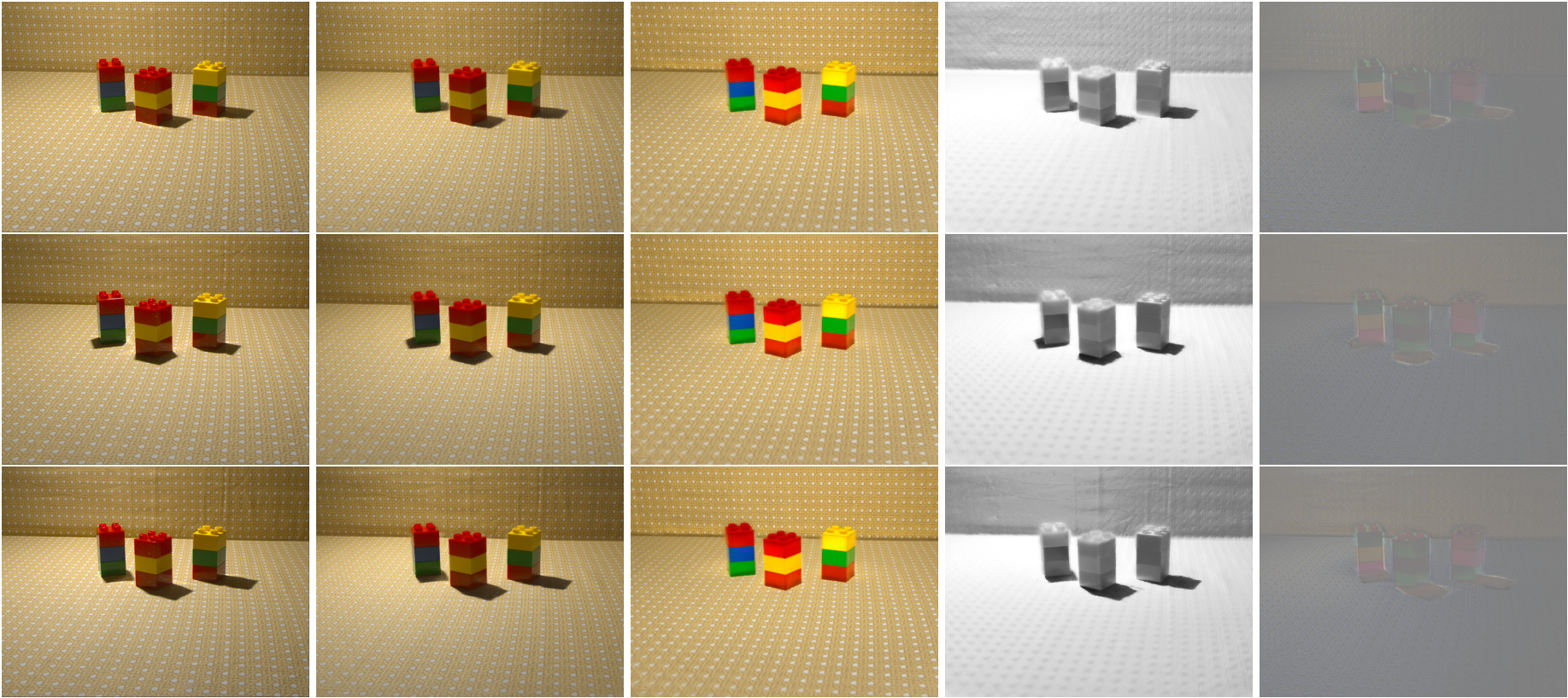}
  \centering
  \begin{minipage}[b]{0.19\linewidth} \centering
   GT  \end{minipage}
   \begin{minipage}[b]{0.19\linewidth} \centering
   Rendering  \end{minipage}
   \begin{minipage}[b]{0.19\linewidth} \centering
   Reflectance  \end{minipage}
   \begin{minipage}[b]{0.19\linewidth} \centering
   Shading  \end{minipage}
   \begin{minipage}[b]{0.19\linewidth} \centering
   Residual  \end{minipage}
  \caption{Qualitative results on the ReNe dataset (Lego). }
  \label{fig: Rene 4}
\end{figure*}

\begin{figure*}[!t]
  \centering
  \includegraphics[width=0.95\textwidth]{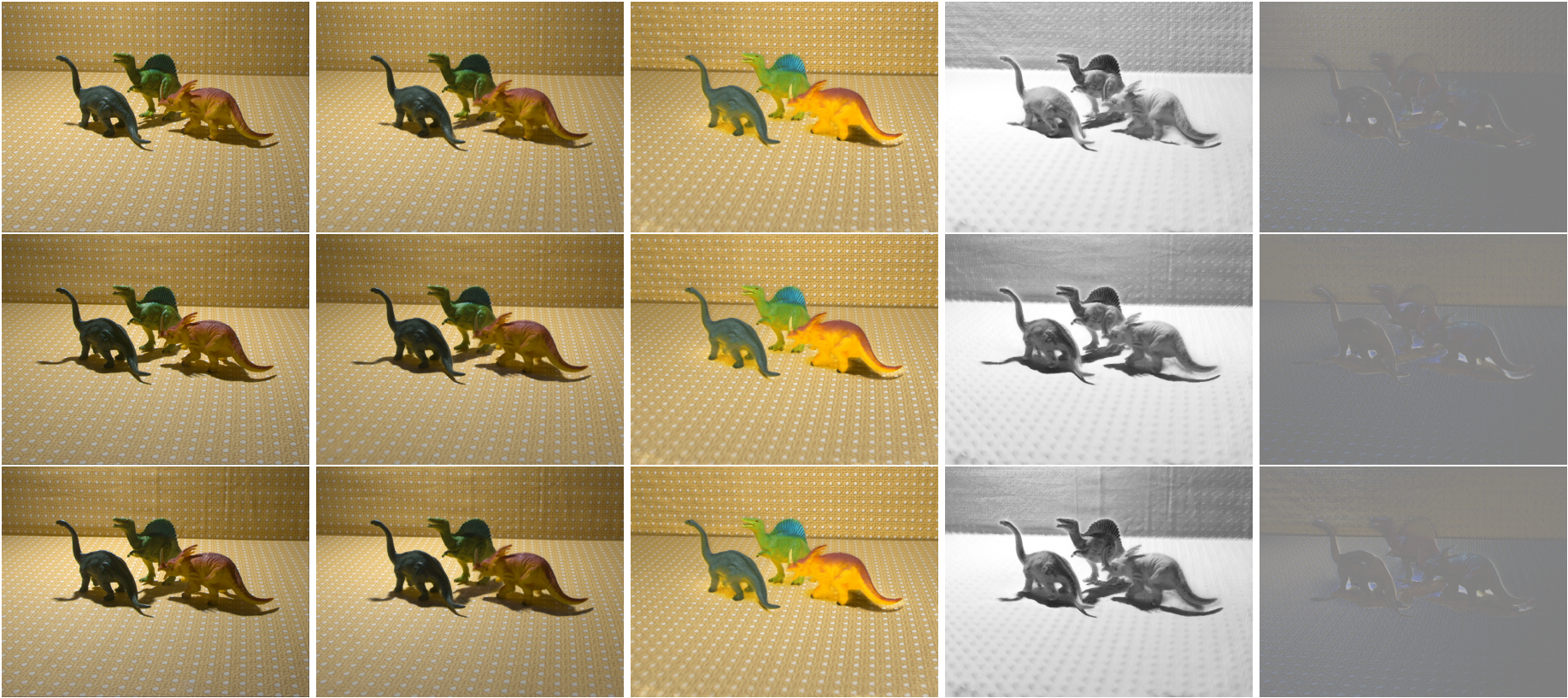}
  \centering
  \begin{minipage}[b]{0.19\linewidth} \centering
   GT  \end{minipage}
   \begin{minipage}[b]{0.19\linewidth} \centering
   Rendering  \end{minipage}
   \begin{minipage}[b]{0.19\linewidth} \centering
   Reflectance  \end{minipage}
   \begin{minipage}[b]{0.19\linewidth} \centering
   Shading  \end{minipage}
   \begin{minipage}[b]{0.19\linewidth} \centering
   Residual  \end{minipage}
  \caption{Qualitative results on the ReNe dataset (Dinosaurs). }
  \label{fig: Rene 5}
\end{figure*}

\begin{figure*}[!t]
  \centering
  \includegraphics[width=0.95\textwidth]{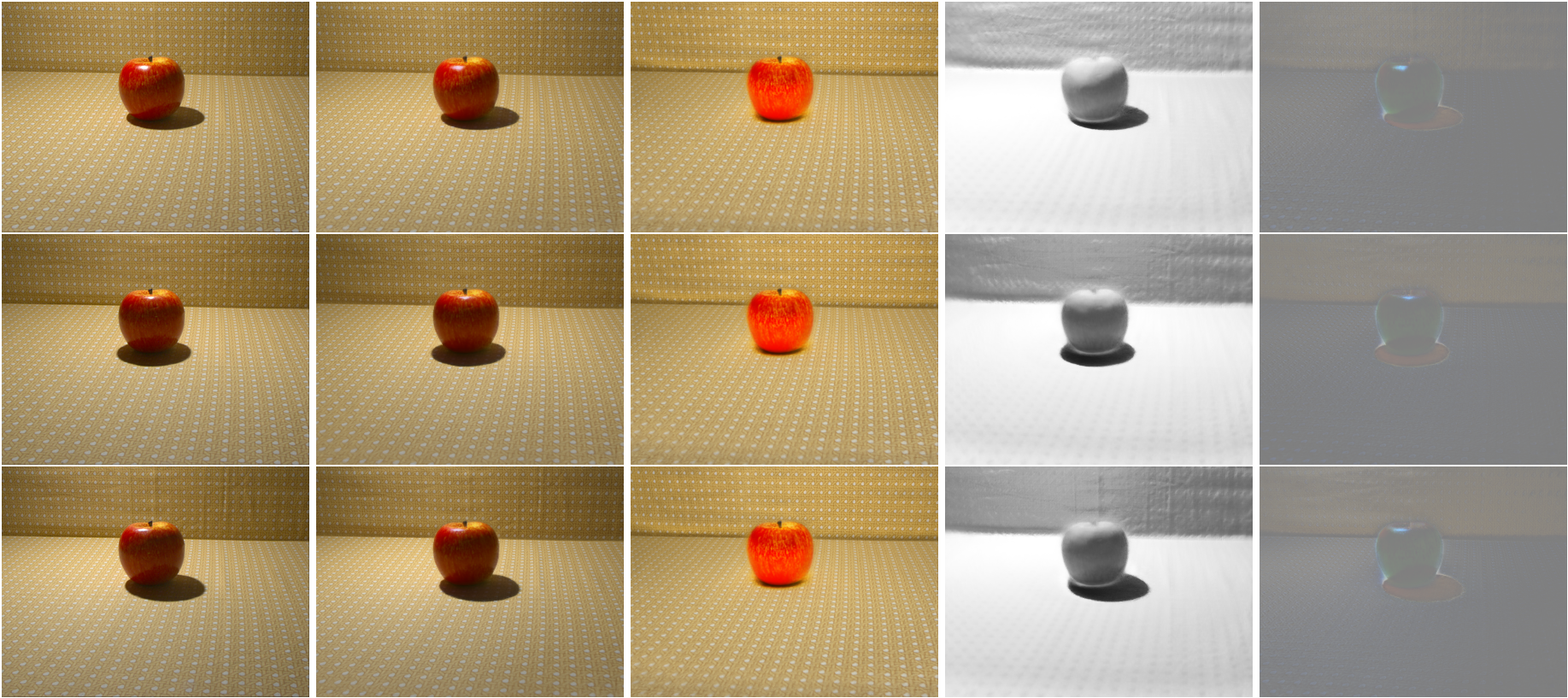}
  \centering
  \begin{minipage}[b]{0.19\linewidth} \centering
   GT  \end{minipage}
   \begin{minipage}[b]{0.19\linewidth} \centering
   Rendering  \end{minipage}
   \begin{minipage}[b]{0.19\linewidth} \centering
   Reflectance  \end{minipage}
   \begin{minipage}[b]{0.19\linewidth} \centering
   Shading  \end{minipage}
   \begin{minipage}[b]{0.19\linewidth} \centering
   Residual  \end{minipage}
  \caption{Qualitative results on the ReNe dataset (Apple). }
  \label{fig: Rene 6}
\end{figure*}

\begin{figure*}[!t]
  \centering
  \includegraphics[width=0.95\textwidth]{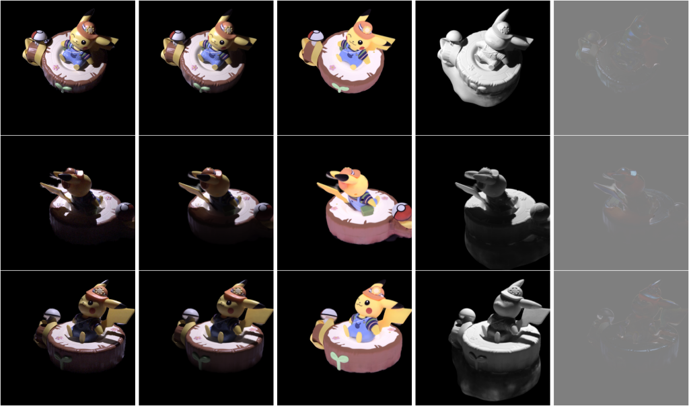}
  \centering
  \begin{minipage}[b]{0.19\linewidth} \centering
   GT  \end{minipage}
   \begin{minipage}[b]{0.19\linewidth} \centering
   Rendering  \end{minipage}
   \begin{minipage}[b]{0.19\linewidth} \centering
   Reflectance  \end{minipage}
   \begin{minipage}[b]{0.19\linewidth} \centering
   Shading  \end{minipage}
   \begin{minipage}[b]{0.19\linewidth} \centering
   Residual  \end{minipage}
  \caption{Qualitative results on the real scene (Pikachu). }
  \label{fig: real 1}
\end{figure*}

\begin{figure*}[!t]
  \centering
  \includegraphics[width=0.95\textwidth]{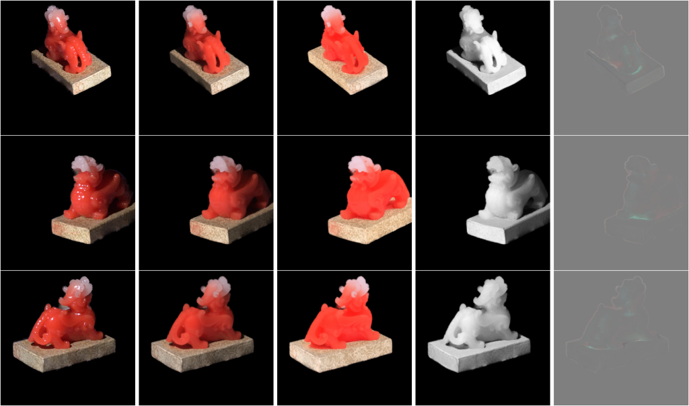}
  \centering
  \begin{minipage}[b]{0.19\linewidth} \centering
   GT  \end{minipage}
   \begin{minipage}[b]{0.19\linewidth} \centering
   Rendering  \end{minipage}
   \begin{minipage}[b]{0.19\linewidth} \centering
   Reflectance  \end{minipage}
   \begin{minipage}[b]{0.19\linewidth} \centering
   Shading  \end{minipage}
   \begin{minipage}[b]{0.19\linewidth} \centering
   Residual  \end{minipage}
  \caption{Qualitative results on the real scene (Pixiu). }
  \label{fig: real 2}
\end{figure*}

\begin{figure*}[!t]
  \centering
  \includegraphics[width=0.95\textwidth]{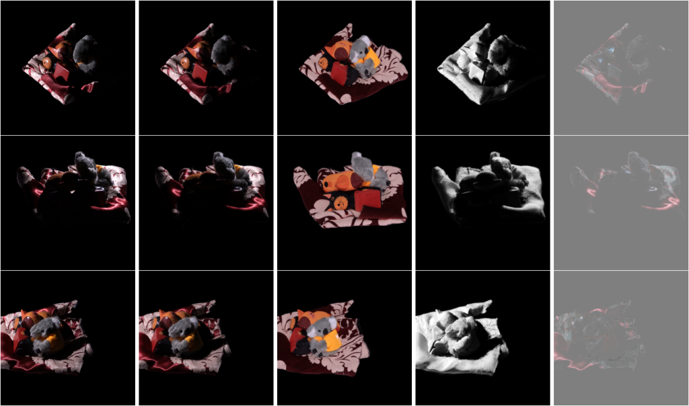}
  \centering
  \begin{minipage}[b]{0.19\linewidth} \centering
   GT  \end{minipage}
   \begin{minipage}[b]{0.19\linewidth} \centering
   Rendering  \end{minipage}
   \begin{minipage}[b]{0.19\linewidth} \centering
   Reflectance  \end{minipage}
   \begin{minipage}[b]{0.19\linewidth} \centering
   Shading  \end{minipage}
   \begin{minipage}[b]{0.19\linewidth} \centering
   Residual  \end{minipage}
  \caption{Qualitative results on the real scene (FurScene). }
  \label{fig: real 3}
\end{figure*}

We present additional results in this section. \cref{Compare with other methods. supplementary} displays additional examples comparing our method with others. \cref{fig: Rene 1} - \cref{fig: Rene 6} shows more qualitative results of our method on the ReNe dataset. Furthermore, \cref{fig: real 1} - \cref{fig: real 3} demonstrate additional qualitative results of our method on real scenes from \cite{zeng2023nrhints}.

